\documentclass[5pt]{article}

\usepackage[letterpaper]{geometry}
\usepackage{spconf,amsmath,epsfig}
\usepackage{enumitem}
\usepackage{multirow, caption}

\usepackage{fancyhdr}
\begin{document}\sloppy

\def\x{{\mathbf x}}
\def\L{{\cal L}}

\title{RAM: A Region-Aware Deep Model for Vehicle Re-Identification}
%
\name{Xiaobin Liu$^1$, Shiliang Zhang$^1$, Qingming Huang$^2$, Wen Gao$^1$}
\address{$^1$School of Electronics Engineering and Computer Science, Peking University, 
China\\
$^2$School of Computer and Control Engineering, University of Chinese Academy of Sciences, 
China\\
\normalsize \{xbliu.vmc, slzhang.jdl, wgao\}@pku.edu.cn, qmhuang@ucas.ac.cn}
%
%
%

\maketitle

\begin{abstract}
Previous works on vehicle Re-ID mainly focus on extracting global features and learning distance metrics. Because some vehicles commonly share same model and maker, it is hard to distinguish them based on their global appearances. Compared with the global appearance, local regions such as decorations and inspection stickers attached to the windshield, may be more distinctive for vehicle Re-ID. To embed the detailed visual cues in those local regions, we propose a Region-Aware deep Model (RAM). Specifically, in addition to extracting global features, RAM also extracts features from a series of local regions. As each local region conveys more distinctive visual cues, RAM encourages the deep model to learn discriminative features. We also introduce a novel learning algorithm to jointly use vehicle IDs, types/models, and colors to train the RAM. This strategy fuses more cues for training and results in more discriminative global and regional features. We evaluate our methods on two large-scale vehicle Re-ID datasets, \emph{i.e.}, \emph{VeRi} and \emph{VehicleID}. Experimental results show our methods achieve promising performance in comparison with recent works.
\end{abstract}
\section{Introduction}
\label{sec:intro}

Vehicle Re-Identification (Re-ID) targets to identify the reappearing vehicles taken by a camera network. It is potential to address the challenging issues like intelligent surveillance video analysis and processing. It is also important for promising applications on intelligent transportation and smart city, such as finding and tracking specific vehicles. Several related tasks on vehicle identification have been extensively studied, such as vehicle attribute prediction~\cite{compcars} and fine-grained vehicle classification~\cite{car196, compcars, vehicleid}. Those tasks mainly focus on identifying the fine-grained categories of vehicles, such as the specific maker and model. Differently, vehicle Re-ID requires the model to distinguish different vehicle instances. As different vehicles of same maker and model may be similar with each other in global appearance, vehicle Re-ID is more challenging and is far from being solved.

\begin{figure}[t]
\begin{center}
\begin{minipage}[b]{1.0\linewidth}
  \centering
  \centerline{\epsfig{figure=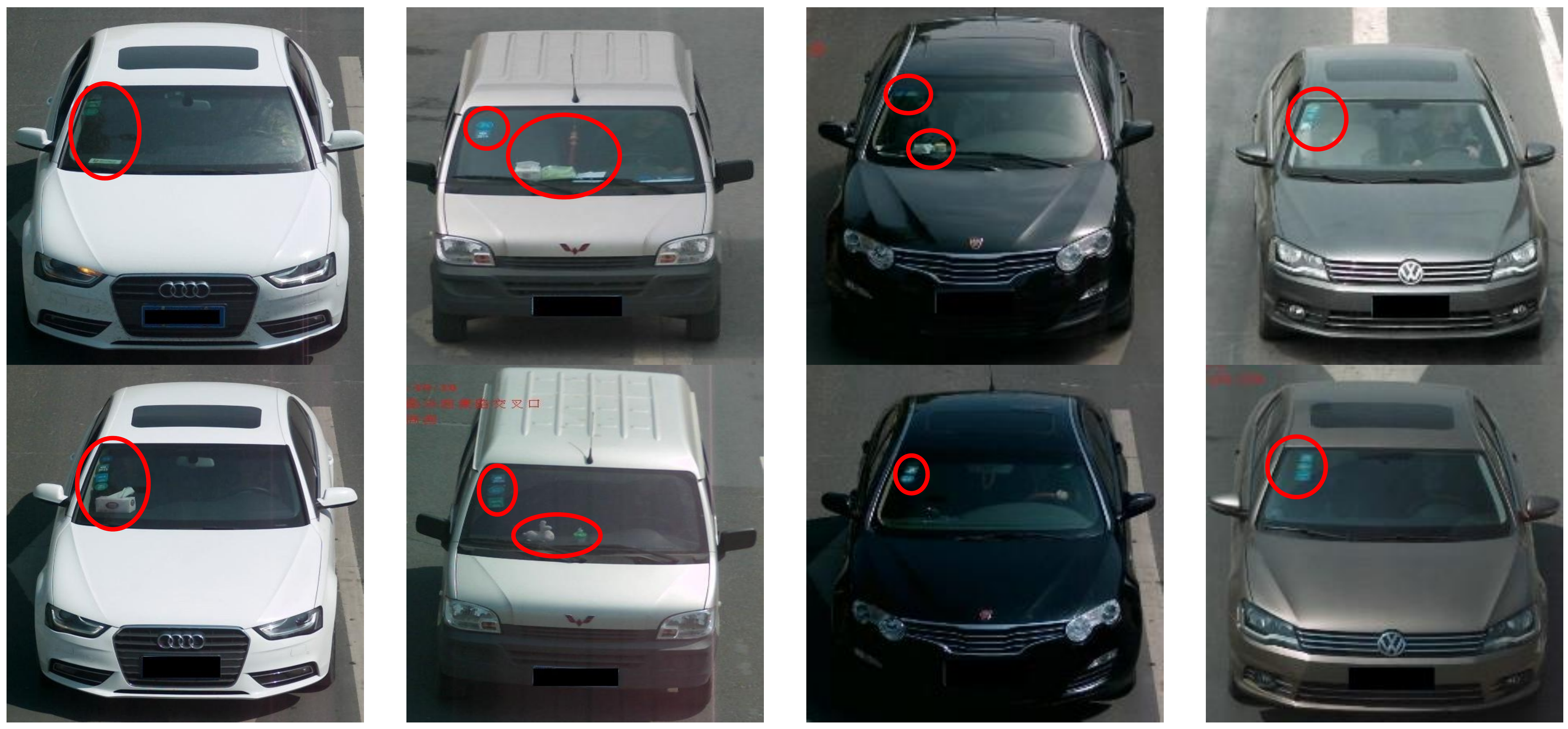, width=8cm}}
\end{minipage}
\end{center}
\caption{Examples of different vehicles with similar global appearance. Each column shows two different vehicles. The differences on local regions are highlighted with red circles. It can be observed that, the differences between similar vehicles mostly lie on some local regions.}
\label{fig:vehicle}
\end{figure}

\begin{figure*}[t]
\begin{center}
\begin{minipage}[b]{1\linewidth}
  \centering
  \centerline{\epsfig{figure=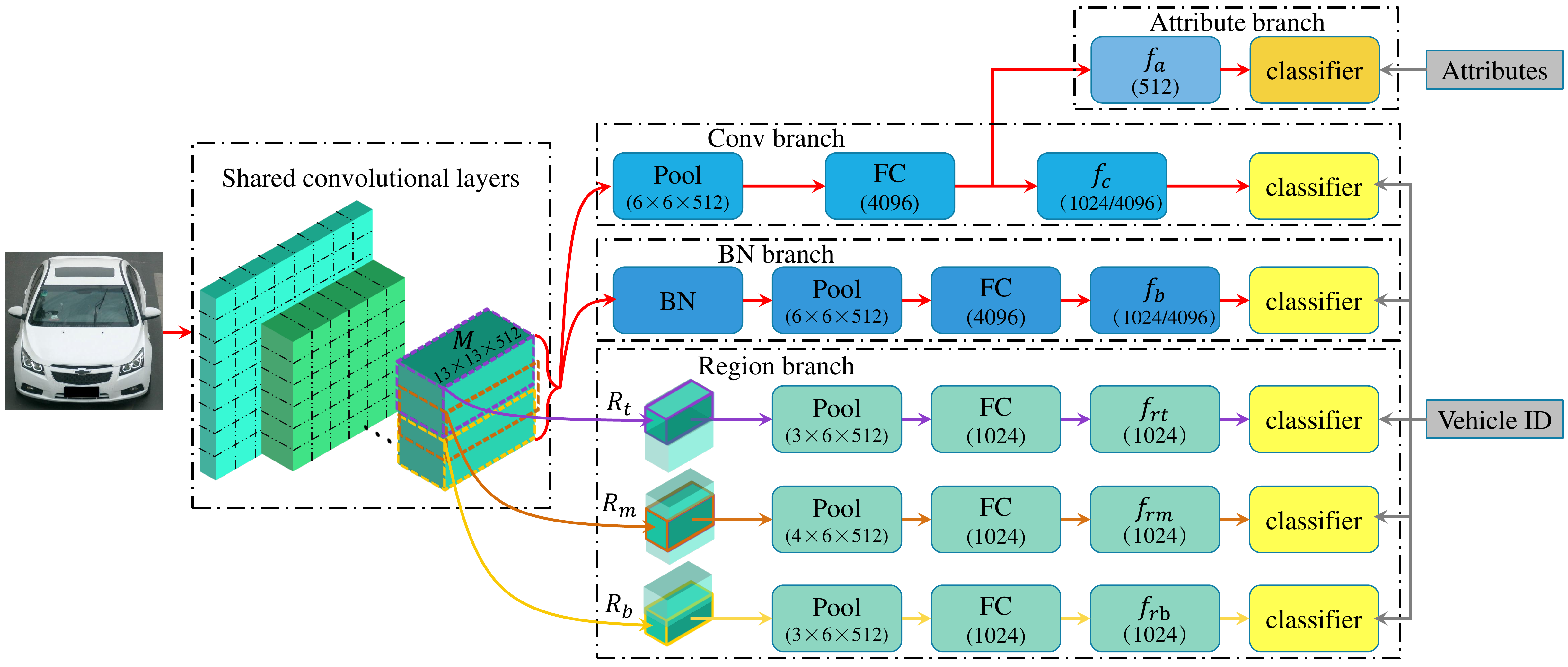, width=15.5cm}}
\end{minipage}
\end{center}
  \vspace{-2mm}
\caption{Structure of the proposed Region-Aware deep Model (RAM), which is composed of several shared convolutional layers and four branches, \emph{i.e.}, Conv branch, BN branch, Regional branch, and Attribute Branch. Each generated feature is trained with an individual classifier. Features are finally concatenated for vehicle instance identification.}
\label{fig:structure}
\vspace{-1mm}
\end{figure*}

Most of existing works focus on designing or learning a robust visual representation for vehicle Re-ID~\cite{veri-eccv, vehicleid}. Some additional clues may also be helpful for this task, \emph{e.g.}, the spatio-temporal information~\cite{veri-eccv,vst,oif}. The quality of surveillance videos is commonly affected by many factors, such as illumination, weather, viewpoint, and occlusion, \emph{etc}. Therefore, hand-crafted features may be unstable in surveillance scenario. Recently, Deep Convolutional Neural Network (DCNN) makes breakthrough in many applications including person Re-ID~\cite{jianing,longhui, longhui2}, fine-grained retrieval~\cite{hantao2, shaoting}, and face recognition~\cite{facenet2015, deepid2}. In those applications, deep features have shown substantial advantages over hand-crafted features in handling noisy visual data. Recently, some datasets have been released to facilitate the research on vehicle Re-ID, such as \emph{VeRi}~\cite{veri-eccv,veri-icme} and \emph{VehicleID}~\cite{vehicleid}. Those large-scale datasets make it possible to train and design DCNN for this task.
Liu \emph{et al.}~\cite{vehicleid} propose a two-branch DCNN structure trained with vehicle IDs and models categories, respectively. They also propose a new distance metric learning method called coupled cluster loss to improve the traditional triplet loss.
Liu \emph{et al.}~\cite{veri-icme} investigate different features and show improvements by fusing hand-crafted features and deep features.
Wang \emph{et al.}~\cite{oif} first predict 20 keypoints and locate 4 regions based on those keypoints. They then fuse local features extracted on those regions and global features for vehicle Re-ID. This work shares certain similarity with ours. However, our method avoids extra annotations and predictions of key points. Moreover, we achieve better performance using a more concise network structure and less training data.

Although previous works have achieved significant progress, vehicle Re-ID still can be improved from many aspects. As most of previous deep learning based works use the global vehicle image as input, their learned descriptions tend to depict the global appearance and may lose discriminative power to local details. As shown in Fig. \ref{fig:vehicle}, different vehicles sharing the same model and maker is similar in global appearance, making it hard to distinguish them. Compared with the global appearance, local regions could be more discriminative. How to effectively extract and embed regional cues for vehicle Re-ID has not been extensively studied. Moreover, fine-grained categorization and vehicle attributes could be helpful in distinguishing vehicles. Although those tasks have been extensively studied in recent years~\cite{car196,compcars}, most of vehicle Re-ID works have not considered to leverage vehicle categories or attributes to help vehicle Re-ID.

To utilize region and attribute cues for vehicle Re-ID, we propose a Region-Aware deep Model (RAM) illustrated in Fig. \ref{fig:structure}. RAM is composed of four branches sharing several convolutional layers. ``Conv branch" learns global features from the whole input image.  ``BN branch" is modified on ``Conv branch" with embedding a Batch Normalization (BN)~\cite{bn} layer to generate complementary global features.  ``Region branch" learns regional features from three overlapping local regions. ``Attribute branch" uses color and model cues to jointly train the model and learn semantic attributes. Features in the four branches are finally concatenated for vehicle Re-ID. As each local region corresponds to a part of the vehicle, RAM encourages the deep model to learn discriminative features for different local regions. RAM also fuses more cues during the training stage and results in more discriminative global and regional features.

We evaluate our methods on two large-scale vehicle Re-ID datasets. Experimental results show that our methods achieve promising performance in comparison with recent works. The contributions of this work can be summarized into two aspects, \emph{i.e.},
\begin{itemize}
\setlength{\itemsep}{0cm}
	\item We propose a Region-Aware deep Model to jointly learn deep features from both the global appearance and local regions. The learned features are more discriminative to detailed local cues on vehicles than previous global ones.
	\item Color and model cues are additionally used to jointly train the deep model. The final concatenated feature achieves promising performance in comparison with recent ones.
\end{itemize}

\section{Proposed Method}
\label{sec:proposed}

Our network structure is illustrated in Fig.~\ref{fig:structure}. Given an input vehicle image, a set of features are generated by RAM. Specifically, five shared convolutional layers generate feature maps $M$. Then, $M$ is fed into four branches to generate different features, \emph{i.e.}, $f_c$ by Conv branch, $f_b$ by BN branch, $f_a$ by Attribute branch, $f_{rt}, f_{rm}$, and $f_{rb}$ by Regional branch from the top, middle and bottom regions of vehicles, respectively. Conv and BN branches generate global feature $f_c$ and $f_b$ from the whole feature maps, respectively. Specially, BN branch adds a Batch Normalization operation~\cite{bn} to the Conv branch to learn complementary global features. Region branch first divides feature maps into three overlapped regions denoted as $R_t$ for top, $R_m$ for middle and $R_b$ for bottom, respectively. And then three sets of full connected layers are used to generate regional features $f_{rt}$, $f_{rm}$ and $f_{rb}$ from corresponding regions. Attribute feature $f_a$ is learned in the Attribute branch. In the following parts, we present the details of the network structure and model training.

\begin{figure}[t]
\begin{minipage}[b]{1.0\linewidth}
  \centering
  \centerline{\epsfig{figure=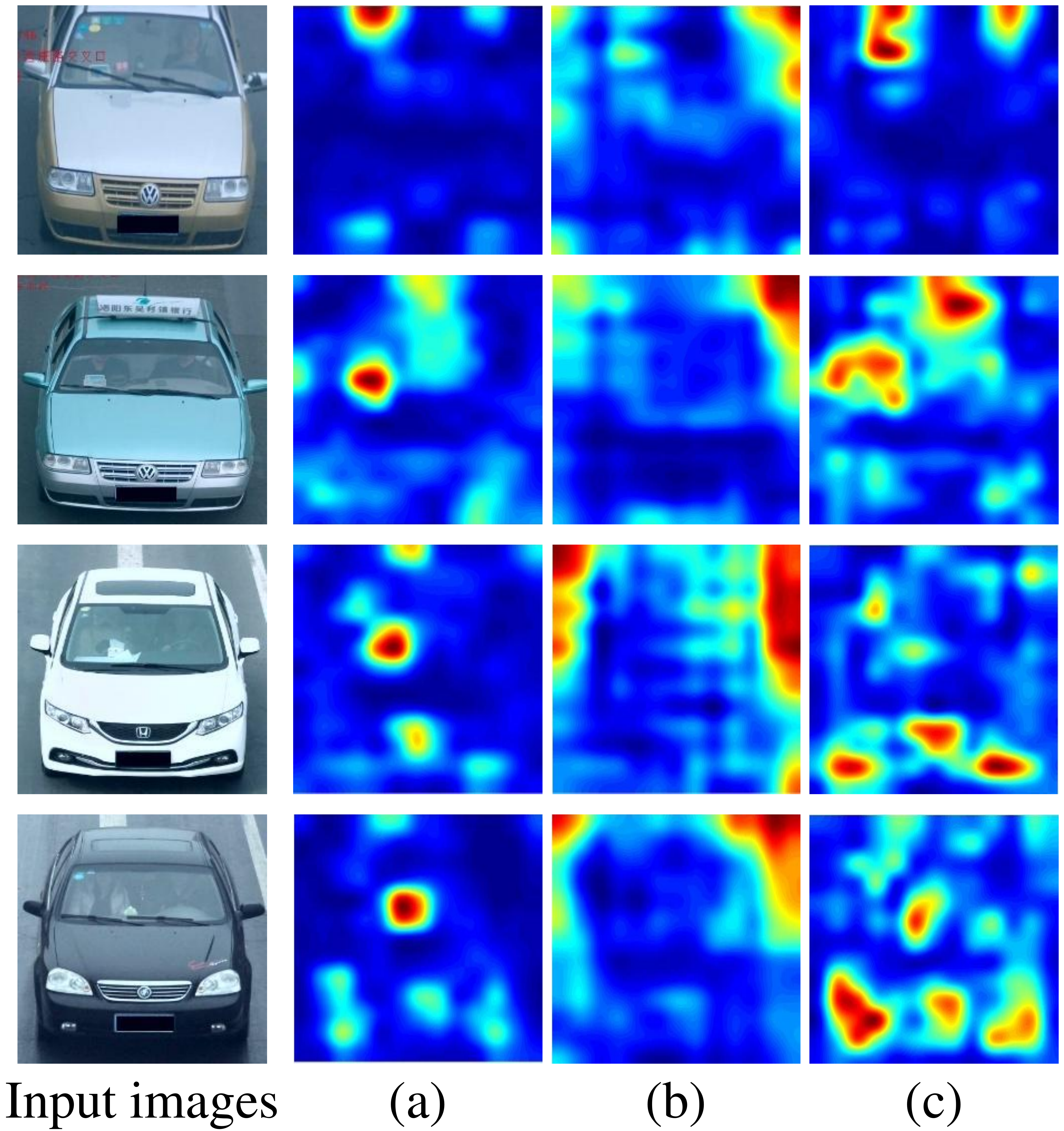, width=7.5cm}}
\end{minipage}
\vspace{-2mm}
\caption{Responses of feature maps generated by different branches. (a), (b), and (c) show the responses of feature maps generated by Conv branch, BN branch, and Region branch, respectively. }
\label{fig:feature_maps}
\vspace{-1mm}
\end{figure}

\subsection{Global Features Extraction}

Conv branch first pools the feature maps $M$ into size $6\times6\times512$ and then uses two Fully Connected (FC) layers to generate feature $f_c$. $f_c$ is trained with vehicle IDs in a classification task. As discussed in~\cite{hantao2}, this network structure and training strategy encourage the network to locate and focus on some regions that are discriminative in vehicle classification. In other words, the local regions effective for minimizing the classification loss will be located. The corresponding feature maps $M$ learned by Conv branch would show higher activation values on those regions. As shown in Fig. \ref{fig:feature_maps} (a), the highly activated regions cover the distinct regions on the vehicles, thus would be important for vehicle classification.

Besides the regions highlighted on $M$, other regions may also be useful for vehicle Re-ID. To make the model focus on more and larger contextual regions, we design a BN branch following Yao \emph{et al.}~\cite{hantao2}, as illustrated in Fig. \ref{fig:structure}. A BN layer is embedded between $M$ and pooling layer to generate new feature maps $M_b$. Then two FC layers are used to generate feature $f_b$. Similarly, a classification task based on vehicle IDs is finally used to train the BN branch.

As discussed in~\cite{hantao2}, BN operation tends to depress the highly activated local regions on the feature maps and increase the visibility of the other regions. This enables the BN branch to depict extra contextual cues in addition to the ones captured by the Conv branch. Some examples of $M_b$ after BN operation are shown in the Fig. \ref{fig:feature_maps} (b). It's clear that $M_b$ depicts larger contextual regions. Fig. \ref{fig:feature_maps} shows that the Conv branch and BN branch depict different cues on the vehicle images, thus may produce complementary global features. We show their complementary in Sec.~\ref{sec:ex_improvements}.

\subsection{Local Features Extraction}

As shown in Fig. \ref{fig:vehicle}, the differences among similar vehicles may lie on some local regions. We hence design a Region branch to generate regional features.

Firstly, Region branch evenly divides feature maps $M$ into $k$ overlapped local regions from top to bottom, as illustrated in Fig. \ref{fig:structure}. We experimentally set $k$ as 3, which could be a reasonable trade-off between network complexity and feature performance.
These local regions are denoted as: $R_t$ for top, $R_m$ for middle and $R_b$ for bottom, respectively. Each of those regions only corresponds to a part of the whole vehicle, \emph{e.g.}, $R_t$ corresponds to car roof and the top of windshield, and $R_b$ may correspond to head lights. We evenly divide the vehicle image because most of the vehicle images are well-aligned. We use the overlapped regions to enhance the robustness of learned features to possible misalignments or viewpoint variations.
A pooling layer is embedded after each region. Then FC layers are applied to generate regional features from each of them, \emph{i.e.} $f_{rt}$ from $R_t$, $f_{rm}$ from $R_m$ and $f_{rb}$ from $R_b$. Finally, a classification task with vehicle ID labels is used to supervise each regional feature learning.

During the training of each branch, FC layers are updated to identify vehicles only with a part of feature maps as input. This procedure enforces the network to extract discriminative details in each region. The regional branch thus has potential to discover more local details than the Conv branch. Some examples of feature maps trained with Region branch are shown in Fig. \ref{fig:feature_maps} (c). It's clear that more discriminative local regions can be identified than the feature maps of Conv branch. It is reasonable to infer that, cues on those local regions will be conveyed in the resulting regional features.

\subsection{Attribute Features Extraction}

Attributes like models, makers, colors, \emph{etc.}, can be regarded as mid-level descriptions to vehicles. Compared with visual features, attributes are more robust to variations of appearance caused by the changes of viewpoints, illuminations, backgrounds, \emph{etc}. Therefore, attribute could be complementary to visual features extracted on global and regional images. We thus use attributes to learn features for vehicle Re-ID.

Attribute prediction can be regarded as an easier identification task than the fine-grained vehicle identification. Therefore, we learn attribute features from the Attribute branch for vehicle Re-ID. As illustrated in Fig.~\ref{fig:structure}, Attribute branch takes the output of the first FC layer in Conv branch as input. Attribute feature $f_a$ is then generated by a FC layer. Finally, attribute feature are learned in an attribute classification task. Compared with directly learning attribute features from input images, this strategy introduces less parameters and makes the training procedure easier.

\subsection{Training}

Every branch in RAM is trained with an individual classification task with softmax loss. The RAM is optimized in multiple classification tasks, and the overall objective function can be formulated as:
\begin{eqnarray}
L(\Theta)=\ell_{conv}+\lambda_1\ell_{BN}+\lambda_2\ell_{re}+\lambda_3\ell_{att},
\label{eqn}
\end{eqnarray}
where $\Theta$ denotes the parameters in the deep model. $\ell_{conv}$, $\ell_{BN}$, $\ell_{re}$, and $\ell_{att}$ denote the classification loss in Conv, BN, Region and Attribute branch, respectively. $\lambda_1$, $\lambda_2$ and $\lambda_3$ denote the weights for corresponding loss. Note that, $\ell_{re}$ is composed of three equally weighted classification losses on different regions.

Training the four branches from scratch could be hard to converge. Instead, we train the model in a step-by-step manner. We first train a model only having the Conv branch. The other branches, \emph{i.e.}, BN, Region and Attribute branches are added orderly. The convolutional layers will be shared by different branches and fine-tuned in multiple classification tasks.

As shown in Fig.~\ref{fig:structure}, the proposed RAM is wide and deep, allowing to utilize multiple supervisions for model training. The resulting features depict vehicles from different aspects. For example, the global and regional features depict the discriminative visual cues. The attribute features depict the attributes and would be more robust to appearance variations and noises. Performance of the learned features will be tested in Sec.~\ref{sec:ex_improvements}.

\section{Experiments}
\label{sec:experiments}

\subsection{Datasets}

We evaluate our method on two large-scale datasets for vehicle Re-ID: \emph{VeRi}~\cite{veri-icme, veri-eccv} and \emph{VehicleID}~\cite{vehicleid}. We first evaluate contributions of each branch in RAM, then make comparison with state-of-the-arts. Details of the two datasets are given as follows:

\emph{VeRi} contains over 50,000 images of 776 vehicles captured by 20 surveillance cameras. Images are annotated with IDs, types and colors. There are 9 types, such as sedan, bus and truck, and 10 colors, such as red, black and orange. 576 vehicles (37,778 images) are used for training and others are used for testing. 1,678 images from the testing set are selected as queries and others are used as gallery images.

\emph{VehicleID} contains 221,763 images of 26,267 vehicles. All of the images are labelled with IDs. A part of vehicles are labelled with colors and vehicle models. 250 models and 7 colors are annotated as attributes. 13,134 vehicles are selected for training and the others are selected for testing. Three subsets are extracted from the testing set with 800, 1600, and 2400 vehicles, respectively. In each subset, an image of each vehicle is randomly selected as gallery image, resulting in 800, 1600, and 2400 gallery images respectively.

\subsection{Implementation Details}

Proposed RAM is implemented with caffe~\cite{caffe2}. We design the structure of RAM based on VGG$\_$CNN$\_$M~\cite{vgg-cnn} and VGG$\_$CNN$\_$M$\_$1024~\cite{vgg-cnn} pre-trained on \emph{ImageNet}~\cite{imagenet} for \emph{VeRi} and \emph{VehicleID}, respectively for fair comparisons with previous methods. All of the loss weights are set to 1 in Eq.~\ref{eqn}. The base learning rate is 0.001 and decreases by multiplying 0.1 after 10 epochs. The size of mini-batch is set to 64.
Images are warped to $224 \times 224$ for both training and testing. Feature maps $M$ is of size $13\times13\times512$. In Region branch, the size of each regions is set as $7\times13\times512$. The size of overlapped region between neighboring regions is $4\times13\times512$.

We use the same evaluation criterion with previous literatures to evaluate the performance. On \emph{VeRi}, the performance is evaluated by mean Average Precision (mAP), Top-1 and Top-5 following~\cite{veri-icme, vst}. On \emph{VehicleID}, Top-1 and Top-5 are used to evaluated the performance following~\cite{vehicleid, oif}.

\subsection{Evaluation of RAM}
\label{sec:ex_improvements}

In this section, we show the improvements gained by each proposed branch in RAM. The evaluation is conducted in four steps:

\emph{Step-1} first trains the \emph{baseline} model only having the Conv branch.

\emph{Step-2} adds the BN branch to the baseline model. Model trained in this step is denoted as \emph{BN}.

\emph{Step-3} further adds the Region branch to model \emph{BN}. Model trained in this step is denoted as \emph{BN+R}.

\emph{Step-4} adds Attribute branch to model \emph{BN+R}. This final model is denoted as \emph{RAM}.

Each step introduces a new branch for feature learning. Therefore, we test the concatenated feature in each step on \emph{VeRi} and \emph{VehicleID}, respectively.

\begin{table}[t]
\begin{center}
\small{
\caption{Performance of features learned by different models on \emph{VeRi}.}
\resizebox{0.4\textwidth}{!}{
\begin{tabular}{c|c|c|c|c}
\hline
 Models 	& Features				& mAP 	& Top-1 & Top-5 \\
\hline
\emph{baseline}		& $f_c$ 					& 0.550 & 0.848	& 0.931\\
\hline
\multirow{3}{*}{\emph{BN}}& $f_c$		& 0.559	& 0.854	& 0.927\\
		& $f_b$						& 0.531	& 0.839	& 0.930\\
     	& $[f_c;f_b]$				& 0.581	& 0.871	& 0.940\\
\hline
\multirow{6}{*}{\emph{BN+R}}	& $f_c$		& 0.556	& 0.852	& 0.925\\
	& $[f_c;f_b]$					& 0.598	& 0.873	& 0.937\\
 	& $[f_c;f_b;f_{rt}]$			& 0.601	& 0.883	& 0.941\\
	& $[f_c;f_b;f_{rm}]$			& 0.590	& 0.874	& 0.934\\
	& $[f_c;f_b;f_{rb}]$			& 0.593	& 0.868	& 0.933\\
	& $[f_c;f_b;f_{r}]$			& 0.609	& \textbf{0.887}	& 0.941\\
\hline
\multirow{4}{*}{\emph{RAM}}&$f_c$		& 0.563	& 0.861	& 0.925\\
	&$[f_c;f_b]$				& 0.604	& 0.869	& 0.937\\
	& $[f_c;f_b;f_{r}]$		& 0.613	& 0.885	& \textbf{0.942}\\
	& $[f_c;f_b;f_{r};f_{a}]$	& \textbf{0.615} & 0.886&0.940\\
\hline

\end{tabular}
}
\label{tab:improvements_veri}
}
\end{center}
\end{table}

\begin{table}[t]
\begin{center}
\caption{Performance of concatenated features learned by different models on \emph{VehicleID}.}
\resizebox{0.48\textwidth}{!}{
\begin{tabular}{c|c|c|c|c|c|c}
\hline
\multirow{2}{*}{{Models}} & \multicolumn{3}{c|}{{Top 1}} & \multicolumn{3}{c}{{Top 5}}\\
\cline{2-7}
			& {Small} 	& {Medium}	& {Large} & {Small} 	& {Medium}	& {Large}\\
\hline

{\emph{baseline}}	 	& {0.694} & {0.673}	& {0.632} & {0.892} & {0.820}	& {0.795}\\

{\emph{BN}}		& {0.722}	& {0.705}	& {0.666} & {0.904}	& {0.853}	& {0.832}\\

{\emph{BN+R}}		& {0.747}	& {0.720}	& {0.674} & {0.908}	& {0.863}	& {0.842} \\

{\emph{RAM}}		& { \textbf{0.752}}	& \textbf{0.723}	& \textbf{0.677} & \textbf{0.915}	& \textbf{0.870}	& \textbf{0.845} \\
\hline
\end{tabular}
}
\label{tab:improvements_vehicleid}
\end{center}
\end{table}

\begin{figure*}[t]
\begin{minipage}[b]{1.0\linewidth}
  \centering
  \centerline{\epsfig{figure=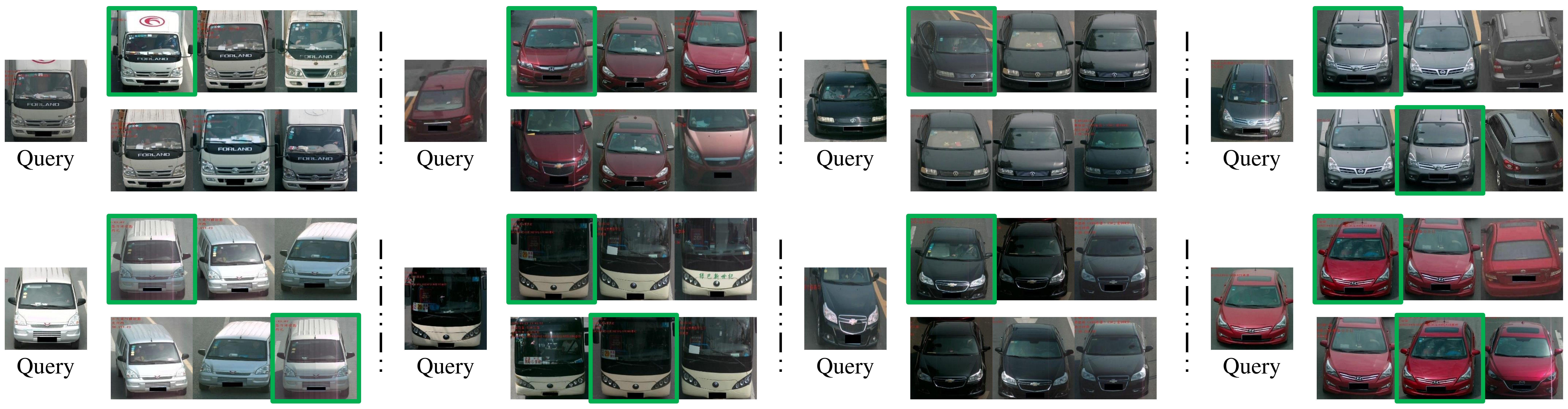, width=17cm}}
\end{minipage}
  \vspace{-5mm}
\caption{Examples of returned images on \emph{VehicleID}. In each example, the first and second row show the top 3 images returned by RAM and baseline model, respectively. True positives are annotated by green bounding boxes.}
  \vspace{-3mm}
\label{fig:return}
\end{figure*}

\subsubsection{Performance on \emph{VeRi}}

The experimental results on \emph{VeRi} are summarized in Table~\ref{tab:improvements_veri}.
Step 1 learns the baseline global feature $f_c$. Step 2 jointly learns features $f_c$ and $f_b$. As shown in Table~\ref{tab:improvements_veri}, $[f_c;f_b]$ performs better than $f_c$ and $f_b$. 
This indicates that $f_b$ extracted by BN branch is an effective complementary to $f_c$. It is also interesting to observe that $f_c$ learned in the \emph{BN} model performs better than the $f_c$ learned in the baseline model. This implies the advantage of network training in multiple tasks.

With \emph{BN+R} model learned in step 3, we test different combinations of regional features. $f_{r}$ denotes the combination of three regional features,\emph{ i.e.}, $f_{r}=[f_{rt};f_{rm};f_{rb}]$. As shown in Table~\ref{tab:improvements_veri}, only fusing single regional feature, \emph{e.g.}, $f_{rb}$ or $f_{rm}$, might degrade the performance because region segmentation could be sensitive to pose variations and misalignment errors. Jointly fusing all regional features, \emph{i.e.}, $[f_c;f_b;f_{r}]$, achieves better performance than the other feature combinations. For example, $[f_c;f_b;f_{r}]$ achieves the mAP of 0.609, significantly better than the 0.601, 0.590, and 0.593 of $[f_c;f_b;f_{rt}]$, $[f_c;f_b;f_{rm}]$, and $[f_c;f_b;f_{rb}]$, respectively. Note that, the concatenated feature $[f_c;f_b]$ of \emph{BN+R} performs better than that of \emph{BN}. This also indicates the advantage of joint feature learning.

With \emph{RAM} learned in the step 4, we compare different concatenations of the resulting features from each branch. Compared with the $[f_c;f_b]$ generated in previous steps, $[f_c;f_b]$ in RAM performs the best. The final feature $[f_c;f_b;f_{r};f_{a}]$ achieves the best performance in mAP among all of the features in Table~\ref{tab:improvements_veri}. It exhibits a substantial improvement of 6.5\% in mAP over the baseline feature.

\subsubsection{Performance on \emph{VehicleID}}

We show the experimental results on \emph{VehicleID} in Table~\ref{tab:improvements_vehicleid}. It's clear that, adding more branches constantly improves the performance. Similar to the observations on \emph{VeRi}, the concatenated feature of \emph{RAM} shows the best Rank-1 and Rank-5 accuracies. Some examples of retrieved images by RAM and baseline are shown in Fig. \ref{fig:return}. In those examples, feature generated by RAM substantially outperforms the baseline global feature. It can be inferred that, RAM is more effective in distinguishing visually similar vehicles. It could also be observed that, RAM is also more robust to viewpoint variations, as shown in the second example of Fig. \ref{fig:return}. Therefore, we can conclude that RAM extracts discriminative features for vehicle Re-ID.

\subsection{Comparison with State-of-the-art Methods}

On \emph{VeRi}, we compare RAM with several recent methods including FACT~\cite{veri-icme}, FACT+Plate-SNN+STR~\cite{veri-eccv}, SCPL~\cite{vst}, OIF~\cite{oif}, and OIF+ST~\cite{oif}. Performance comparison is summarized in Table~\ref{tab:comp_veri}.
On \emph{VehicleID}, the compared works includes VGG+T~\cite{vehicleid}, VGG+CCL~\cite{vehicleid}, Mixed Diff+CCL~\cite{vehicleid} and OIF~\cite{oif}, as shown in Table~\ref{tab:comp_vehicleid}. In Table~\ref{tab:comp_veri} and Table~\ref{tab:comp_vehicleid}, we use ``*" to denote the model that uses ResNet~\cite{resnet} as base network, which is deeper than our base network VGG$\_$CNN$\_$M. We use superscript ``+" to denote methods that are trained on a bigger training set, which is merged by the training data of \emph{VeRi}, \emph{VehicleID}, \emph{CompCars}~\cite{compcars} and \emph{BoxCars21K}~\cite{boxcars}.

It can be observed that, RAM achieves the best performance on two datasets compared with others.
Compared with SCPL~\cite{vst}, which uses a deeper network and extra spatial-temporal cues, RAM outperforms it by 3.2\% in mAP on \emph{VeRi}. Compared with OIF~\cite{oif} that uses more training data and a more complex structure, RAM outperforms it by 13.5\% in mAP on \emph{VeRi} and by 0.7\% in Top-1 accuracy on \emph{VehicleID}.

\begin{table}[t]
\begin{center}
\caption{Comparison with recent works on \emph{VeRi}.}
\resizebox{0.4\textwidth}{!}{
\begin{tabular}{c|c|c|c}
\hline
Method									& mAP 		& Top-1 & Top-5 \\
\hline
FACT~\cite{veri-icme} 					& 0.199 	& 0.597	& 0.753\\
FACT+Plate-SNN+STR~\cite{veri-eccv}		& 0.278		& 0.614	& 0.788\\
SCPL$^*$~\cite{vst} 					& 0.583 	& 0.835 & 0.900\\
OIF$^+$~\cite{oif} 						& 0.480 	& 0.659 & 0.877\\
OIF+ST$^+$~\cite{oif} 					& 0.514 	& 0.683 & 0.897\\
\hline
RAM					& \textbf{0.615}	& \textbf{0.886}	& \textbf{0.940}\\
\hline
\end{tabular}
}
\label{tab:comp_veri}
\vspace{-4mm}
\end{center}
\end{table}

\begin{table}[t]
\begin{center}
\caption{Comparison with recent works on \emph{VehicleID}.}
\resizebox{0.48\textwidth}{!}{
\begin{tabular}{c|c|c|c|c|c|c}

\hline

\multirow{2}{*}{Method}		& \multicolumn{3}{c|}{Top 1} & \multicolumn{3}{c}{Top 5} \\
\cline{2-7}

	& Small & Medium & Large & Small & Medium & Large \\
\hline

VGG+T~\cite{vehicleid} & 0.404 & 0.354 & 0.319 & 0.617 & 0.546 & 0.503\\

VGG+CCL~\cite{vehicleid} & 0.436 & 0.370 & 0.329 & 0.642 & 0.571 & 0.533\\

Mixed Diff+CCL~\cite{vehicleid} & 0.490 & 0.428 & 0.382 & 0.735 & 0.668 & 0.616\\

OIF$^+$~\cite{oif} 	& - & - & 0.670 & - & - & 0.829 \\

\hline

RAM	& \textbf{0.752}	& \textbf{0.723}	& \textbf{0.677} & \textbf{0.915}	& \textbf{0.870}	& \textbf{0.845}\\

\hline

\end{tabular}
}
\vspace{-0.4cm}
\label{tab:comp_vehicleid}
\end{center}
\end{table}

\section{Conclusion}
\label{sec:conclusion}
This paper presents a Region-Aware deep Model (RAM) for vehicle Re-ID task. In addition to global features, RAM includes a Region branch to extract regional features from three overlapped local regions. This encourages the deep model to focus on more details in local regions and results in more discriminative features. We also jointly train an Attribute branch to generate attribute features, which are potential to be more robust to viewpoint variations. Experiments on two large-scale vehicle datasets demonstrate that RAM extracts discriminative features and achieves promising performance.

\vspace{3mm}
\textbf{Acknowledgements}
This work is supported by National Science Foundation of China under Grant No. 61572050, 91538111, 61620106009, 61429201, and the National 1000 Youth Talents Plan, in part to Dr. Qi Tian by ARO grant W911NF-15-1-0290 and Faculty Research Gift Awards by NEC Laboratories of America and Blippar.

\ninept
\bibliographystyle{IEEEbib}
\bibliography{camera-ready_icme2018template}

\end{document}